%% file: 0main.tex
\PassOptionsToPackage{table,dvipsnames}{xcolor}

\documentclass[sn-mathphys-num]{sn-jnl}% Math and Physical Sciences Numbered Reference Style 

\usepackage{amsmath,amssymb,amsfonts,bbm}
\usepackage{graphicx}
\usepackage{booktabs,multirow,array,tabularx}
\usepackage{xspace}
\usepackage[table]{xcolor}
\usepackage{tikz}
\usetikzlibrary{shapes.geometric, backgrounds, positioning, fit, arrows.meta}
\usepackage{pgf-pie}
\usepackage{subcaption}
\usepackage[utf8]{inputenc}
\usepackage{tcolorbox}
\usepackage{listings}
\usepackage{booktabs}
\usepackage{array}
\usepackage{tabularx}
\usepackage{adjustbox}  % safer than resizebox

\usepackage{threeparttable}
\usepackage{xurl} % better URL breaking
\usepackage{lipsum}

\newcommand{\huggingface}{\raisebox{-1.5pt}{\includegraphics[height=2ex]{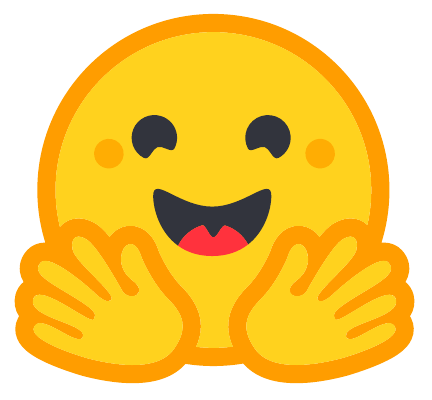}}\xspace}
\newcommand{\git}{\raisebox{0pt}{\includegraphics[height=2ex]{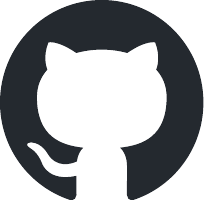}}\xspace}

\begin{document}
\setcounter{page}{1} 

\title{Responsible AI in NLP: \textbf{GUS-Net} Span-Level Bias Detection Dataset and Benchmark for Generalizations, Unfairness, and Stereotypes}

\author*[1]{\fnm{Maximus} \sur{Powers}}\email{powersms@clarksonalumni.com}\equalcont{These authors contributed equally.}
\author[2]{\fnm{Shaina} \sur{Raza}}\email{shaina.raza@torontomu.ca}\equalcont{These authors contributed equally.}
\author[3]{\fnm{Alex} \sur{Chang}}\email{tchang54@asu.edu}
\author[4]{\fnm{Rehana} \sur{Riaz}}\email{rehanaNoorani3@gmail.com}
\author[3]{\fnm{Umang} \sur{Mavani}}\email{umavani@asu.edu}
\author[3]{\fnm{Harshitha Reddy} \sur{Jonala}}\email{hjonnal1@asu.edu}
\author[3]{\fnm{Ansh} \sur{Tiwari}}\email{atiwar31@asu.edu}
\author[3]{\fnm{Hua} \sur{Wei}}\email{hua.wei@asu.edu}

\affil[1]{\orgname{Clarkson University}}
\affil[2]{\orgname{Toronto Metropolitan University}}
\affil[3]{\orgname{Arizona State University}}
\affil[4]{\orgname{Independent Researcher}}

\abstract{
Representational harms in language technologies often occur in short spans within otherwise neutral text, where phrases may simultaneously convey generalizations, unfairness, or stereotypes. Framing bias detection as sentence-level classification obscures which words carry bias and what type is present, limiting both auditability and targeted mitigation. We introduce the \textbf{GUS-Net Framework}, comprising the \textbf{GUS dataset} and a \textbf{multi-label token-level detector} for span-level analysis of social bias. The GUS dataset contains 3,739 unique snippets across multiple domains, with over \textbf{69,000 token-level annotations}. Each token is labeled using BIO tags (Begin, Inside, Outside) for three pathways of representational harm: \textbf{Generalizations, Unfairness,} and \textbf{Stereotypes}. To ensure reliable data annotation, we employ an automated multi-agent pipeline that proposes candidate spans which are subsequently verified and corrected by human experts. We formulate bias detection as multi-label token-level classification and benchmark both encoder-based models (e.g., BERT family variants) and decoder-based large language models (LLMs). Our evaluations cover token-level identification and span-level entity recognition on our test set, and out-of-distribution generalization. Empirical results show that encoder-based models consistently outperform decoder-based baselines on nuanced and overlapping spans while being more computationally efficient. The framework delivers interpretable, fine-grained diagnostics that enable systematic auditing and mitigation of representational harms in real-world NLP systems.
}


\maketitle

%% --- Optional: move HF/GitHub links here or in Declarations ---
\begin{center}
\begin{tabular}{cll}
\huggingface & \textbf{Collection:} & {\small\href{https://huggingface.co/collections/maximuspowers/gus-net-social-bias-ner-685f3eb04a8c3fb9e706cb96}{\nolinkurl{GUS-Net}}}\\
\git & \textbf{Code:} & {\small\href{https://github.com/maximus-powers/gus-net}{\nolinkurl{GUS-Net}}}
\end{tabular}
\end{center}

%% --- Main sections ---
\input{1intro}
\input{2related-works}

\input{3method}
\input{4experiment}
\input{4Results}

\input{5conclusion}

%% --- References ---
\bibliography{references}

\appendix
\input{appendix}

\end{document}

%% file: 1intro.tex
\section{Introduction}
\label{sec:intro}

The widespread adoption of large language models (LLMs) and Small Language Models (SLMs) \cite{thawakar2024mobillama} in domains such as education, healthcare, and recruitment has intensified concern about their tendency to absorb and amplify existing social biases \cite{kumar2024decoding}. Explicitly derogatory language is relatively easy to detect, but implicit bias is harder: it appears as context-dependent expectations or seemingly benign statements that nonetheless reinforce stereotypes \cite{raza2024developing}. For example, praising a woman CEO for balancing work and family inadvertently assumes that such a burden is uniquely feminine (Fig.\ref{fig:covert_bias}). Responsible Natuaral Language Processing (NLP) research therefore requires methods that surface subtle bias and give users actionable explanations.

\begin{figure}[h]
\centering
\begin{tcolorbox}[colback=gray!10,colframe=black!40,
  width=0.9\linewidth,boxrule=0.5pt,arc=2pt,auto outer arc]
\textbf{Example of Covert Bias in Language}\\[4pt]
\textit{``The female CEO balanced her work and family commitments admirably, which some described as surprising.''}
% \\[6pt]
\end{tcolorbox}
\caption{Illustration of covert bias in a seemingly positive statement. Although the sentence appears complimentary, it subtly reinforces a stereotype. The noun phrase ``female CEO'' marks gender as salient for leadership, and the word ``surprising'' presupposes that such success is unexpected for a woman leader. Together they encode a gendered double standard that is rarely applied to men. This illustrates how praise can perpetuate stereotypes and motivates careful, span level analysis in responsible NLP research~\cite{jacobs2020meaning}.}
\label{fig:covert_bias}
\end{figure}

Concerns about social bias in NLP span two types of harm. Allocational harms occur when systems distribute resources or opportunities unequally across groups \cite{cyberey-etal-2025-prevalent, qureshi2025thinking}. Representational harms arise when language disparages, stereotypes, or erases groups by reinforcing negative associations \cite{shahbazi2023representation, narnaware2025sb}. In this work, we focus on representational harms in text. We categorize biased spans into three distinct pathways: (i) assigning attributes to a group (stereotypes), (ii) attaching unfair or pejorative descriptors (unfairness), and (iii) making categorical or quantificational claims about groups (generalizations). These categories are informed by sociolinguistic theory and prior bias scholarship \cite{hovy2021five}. Although these bias types often co-occur, most existing approaches collapse them into a single binary sentence-level label \cite{dixon2018measuring,razaCBDT,tokpo_how_2023}, which obscures the specific words carrying bias and the type of harm involved. This limits interpretability and targeted mitigation. In this work, we address this gap by pursuing span-level detection of bias type, enabling overlapping categories to be explicitly disambiguated.

\begin{figure*}[t]
\centering
\includegraphics[width=0.95\textwidth]{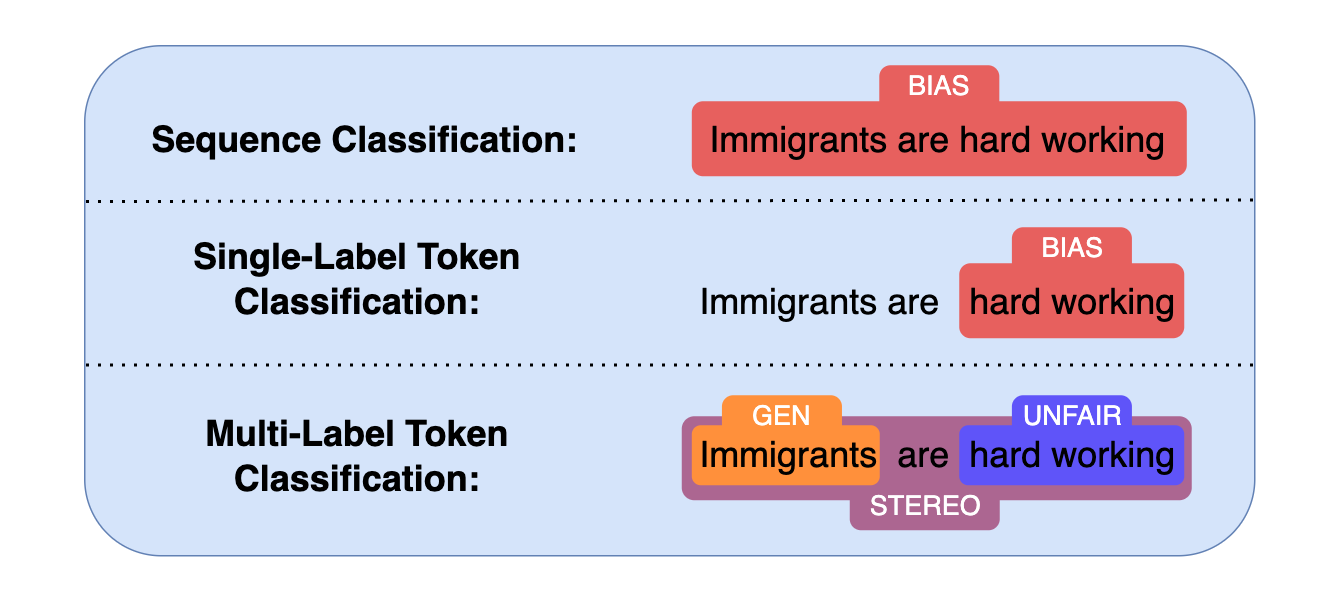}  \caption{Traditional sequence and token classification tasks compared with the proposed multi label token classification approach.}
\label{fig:various-tasks}  
\end{figure*}

As mentioned earlier, most prior work casts bias detection as sentence level binary or multiclass classification \cite{jigsawtoxiccomment,dixon2018measuring}. Transformer based encoders such as BERT \cite{devlin2018bert}, RoBERTa \cite{liu2019robertarobustlyoptimizedbert}, and DeBERTa \cite{he2020deberta}, as well as generative LLMs, have been applied to this problem \cite{raza2025fake}. However, sentence level labels hide which words or phrases carry bias and which type of bias is present \cite{vayani2025all, raza2025vldbench}. Recently, token level approaches inspired by named entity recognition (NER) have emerged \cite{sharnagat2014named}. Datasets such as NBIAS \cite{raza2024nbias} and the Media Bias Annotation Corpus \cite{spinde2021mbicmediabias} provide single “BIAS” tags at the token level, sometimes with coarse domain labels such as media versus news. BABE \cite{spinde2022neural} extends this line of work with expert annotation but still treats all biased tokens uniformly and remains restricted to the news domain. As illustrated in Figure~\ref{fig:various-tasks}, sequence classification assigns a single global label, single label token classification collapses distinct loci of bias into one tag, whereas a multi label token formulation exposes which words specifically carry generalizations, unfairness, or stereotypes. Consequently, the field lacks a large scale, multi domain corpus with multi label span annotations that distinguish generalizations, unfairness, and stereotypes.

We introduce the \textbf{Generalizations, Unfairness, and Stereotypes (GUS) }-Net framework. At its core is the GUS dataset, comprising 3,739 text snippets with over 69,000 token-level annotations for bias entities across diverse domains, including religion, race, gender, politics, nationality, and age. Each token is annotated with one or more of four BIO-tagged labels: B/I-GEN, B/I-UNFAIR, B/I-STEREO, and O (neutral). The BIO scheme: “Begin, Inside, Outside”, marks the first token of a labeled span, the subsequent tokens within that span, and tokens outside any span \cite{sharnagat2014named, perera2020named}. Because bias spans may overlap, a token can receive multiple labels, which our framework accommodates. To construct the corpus, we synthesized candidate sentences through controllable LLM prompting and applied a two-step annotation process: an agent first generated candidate spans, which were then verified and corrected by human experts. This hybrid pipeline provides broader coverage than prior datasets, which are often limited to news text or single-label annotations.

On top of the data, we formulate bias detection as a \textit{multi-label token classification} task and evaluate both encoder-based discriminative models and decoder-based generative models underBIO alignment. We use focal loss \cite{lin2018focallossdenseobject} and threshold tuning to handle the extreme class imbalance (the majority of tokens are neutral) and benchmark performance with token-level precision, recall, F1, macro-F1 and Hamming loss. We additionally test out-of-distribution generalization on BABE dataset \cite{spinde2022neural}by correlating normalized positive-tag rates with BABE expert-annotated bias densities. Finally, we perform ablation studies (replacing GUS with re-annotated BABE and substituting focal loss with binary cross-entropy) and hyperparameter sweeps for focal-loss parameters $\alpha$ and $\gamma$.

\textbf{Contributions.} 
Our key contributions are as:
\begin{itemize}
    \item We construct a new multi-domain corpus with 3,739 sentences with over 69,000 token-level annotations at the token level for \emph{Generalizations}, \emph{Unfairness}, \emph{Stereotypes}, and \emph{Neutral}. Our hybrid LLM–human pipeline scales annotation while maintaining quality.
    \item We cast bias detection as a multi-label BIO tagging problem and release the GUS annotation guidelines and evaluation scripts to the community.
    \item We benchmark state-of-the-art encoder-only models (BERT, DistilBERT, RoBERTa) and decoder-only LLMs (instruction-tuned and few-shot) under strict alignment, showing that encoders achieve higher macro F1 and lower Hamming loss while generative models suffer from output-length and alignment issues.
    \item We provide out-of-distribution validation on the BABE corpus, showing positive correlations between GUS-Net’s normalized bias density and expert-annotated bias density, demonstrating transferability beyond the training domain.
    \item We perform ablation studies and sensitivity analysis, demonstrating that both the GUS dataset and focal loss are essential for strong performance and that the chosen focal-loss parameters $(\alpha=0.65,\gamma=2)$ yield a stable operating point.
\end{itemize}
\noindent
Together, these contributions advance the study of fine-grained bias detection by providing the first large-scale, multi-label, token-level corpus and demonstrating the efficacy of encoder-only models for structured bias tagging. Our experiments highlight both the promise of multi-label token classification and the limitations of existing generative approaches, guiding future work in this important area.

%% file: 2related-works.tex
\section{Related Work}
\label{sec:related}

\subsection{Bias in Large Language Models}
Recent advances in LLMs have intensified concerns around social bias in NLP systems. LLMs are trained on vast web corpora that encode historical stereotypes and social inequalities \cite{zhao2023survey, kasneci2023chatgpt, da2024open}. Empirical studies have shown that such models perpetuate occupational gender stereotypes \cite{heilman2012gender}, racial sentiment skew \cite{jin2023darkbert}, and age-based biases \cite{raza2024dbias}. As a result, surveys such as \cite{hovy2021five} and \cite{sun-etal-2019-mitigating} advocate for fairness-aware evaluation protocols that go beyond simple performance metrics and instead focus on disparate impacts across demographic groups.

\begin{table*}[h]
\centering
\caption{Summary of representative datasets and benchmarks for bias detection and evaluation in NLP. Sentence-level (sent.) and token-level (token) granularity are indicated.}
\resizebox{\textwidth}{!}{
\begin{tabular}{lcccccl}
\toprule
\textbf{Dataset / Benchmark} & \textbf{Year} & \textbf{Granularity} & \textbf{Labels} & \textbf{Size} & \textbf{Domain} & \textbf{Notes} \\
\midrule
MBIC \cite{spinde2021mbicmediabias} & 2021 & Token + Sentence & BIAS & 1{,}700 & News & Annotator metadata \\
BABE \cite{spinde2022neural} & 2022 & Token + Sentence & BIAS & 3{,}700 & News & Expert annotators \\
Toxic Spans \cite{pavlopoulos-etal-2021-semeval} & 2021 & Token spans & Toxic & $\sim$10{,}000 & Comments & SemEval shared task \\
N-BIAS \cite{raza2024nbias} & 2024 & Token & BIAS & 3{,}700 & News/Tweets & BIO tagging scheme \\
BOLD \cite{dhamala2021bold} & 2022 & Sentence & Demographic axes & 23{,}000 & Multi-modal & 14 languages + vision \\
CrowS-Pairs \cite{nangia2020crows} & 2020 & Sentence pairs & Stereotypes & 1{,}500 & English & Contrastive pairs \\
HolisticBias \cite{smith2022m} & 2022 & Sentence & 13{,}000 identities & 113{,}000 & English & Identity coverage \\
BEADs \cite{raza2024beads} & 2024 & Multi-task & Bias axes & 50{,}000+ & Multi-domain & GPT-4 + human validation \\
SHADES \cite{mitchell2025shades} & 2025 & Sentence & Stereotypes & $\sim$4{,}800 & 16 languages & Cultural bias analysis \\
Anno-Lexical \cite{horych2024annolexical} & 2025 & Sentence & Lexical bias & 48{,}000 & Multi-domain & LLM + human hybrid pipeline \\ \hline
\textbf{GUS (ours)} & 2025 & Token & GEN, UNFAIR, STEREO & 3{,}700 & Multi-domain & Multi-label spans \\
\bottomrule
\end{tabular}
}

\label{tab:bias-datasets}
\end{table*}

\subsection{Span-Level Bias Corpora}
While many early approaches to bias detection operate at the sentence or document level, recent work has explored fine-grained, span-level annotations to identify specific biased expressions. The MBIC dataset provides token- and sentence-level annotations over 1{,}700 media statements labeled with the entity \textit{BIAS} \cite{spinde2021mbicmediabias}, while BABE extends this work with expert-labeled bias spans across 3{,}700 U.S. news sentences \cite{spinde2022neural}. The SemEval 2021 Toxic Spans task further demonstrated the feasibility of subjective token labeling at scale, releasing approximately 10{,}000 crowd-annotated online comments \cite{pavlopoulos-etal-2021-semeval}. More recent efforts include NBIAS, which reframes bias detection as a BIO tagging task \cite{raza2024nbias}.

\subsection{Annotation Strategies}
The development of token-level bias datasets is inherently resource-intensive. To alleviate this, hybrid annotation frameworks have emerged that combine LLM-generated labels with human verification. For instance, the Anno-Lexical corpus offers 48{,}000 synthetically labeled sentences generated through prompting and validated through expert review \cite{horych2024annolexical}. Other strategies use multi-agent pipelines to decompose statements into fact–opinion structures before assigning bias scores \cite{da2024prompt}. Although these approaches increase scalability, they still require expert oversight to reliably detect subtle and context-sensitive forms of bias \cite{vidgof2023large}.

\subsection{Cross-Modal Benchmarks}
Bias manifests differently across linguistic and cultural contexts, motivating the development of multilingual benchmarks. The BOLD benchmark measures demographic bias across fourteen languages in tasks ranging from image captioning to dialogue \cite{dhamala2021bold}. CrowS-Pairs \cite{nangia2020crows} and StereoSet \cite{nadeem2020stereoset} introduce minimal-pair sentence structures to reveal stereotypical preferences, while HolisticBias \cite{smith2022m} expands coverage to 13{,}000 identity descriptors. The BEADs benchmark incorporates over 50{,}000 examples across classification, generation, and token labeling tasks, with GPT-4 responses verified by experts \cite{raza2024beads}. Similarly, SHADES introduces over 300 stereotype scenarios translated into sixteen languages to evaluate cultural sensitivity in LLMs \cite{mitchell2025shades}.

\subsection{Modeling Approaches}
Traditionally, bias detection is commonly treated as binary or multiclass text classification task where aim is to assign labels such as “biased” or “neutral” to text segment like sentence, paragraph or documents. Early approaches to this problem often employed encoder-only transformer architectures such as BERT \cite{devlin2018bert}, RoBERTa \cite{liu2019robertarobustlyoptimizedbert}, or DeBERTa \cite{he2020deberta}, which demonstrated strong performance on standard classification benchmarks \cite{dixon2018measuring, li2020survey}. These models are typically fine-tuned on task-specific datasets and are well-suited for structured inputs and outputs.

More recently, the focus has shifted toward token-level bias detection, where the goal is to identify and label specific words or spans within a text that convey biased sentiment or framing. Token classification tasks often adopt BIO (Begin–Inside–Outside) tagging schemes, similar to those used in named entity recognition, enabling models to capture multi-token expressions and their semantic structure \cite{raza2024nbias}. 
Generative LLMs, such as GPT-3.5, GPT-4, and their instruction-tuned variants, have also been explored for bias detection \cite{kumar2024decoding,raza2024developing,raza2023constructing}. These models are evaluated via few-shot or zero-shot prompting, where the model is given a short set of examples or an instruction and asked to classify or explain bias within new input texts \cite{wu2023rethinking, zhao2023survey}. While generative models excel at producing nuanced, human-like rationales and supporting multi-turn interactions, their token-level precision often lags behind that of encoder-only models. Furthermore, generative models are susceptible to hallucinations and may reflect biases embedded in their pretraining data unless specifically aligned or fine-tuned to counteract them \cite{vidgof2023large}.

Despite progress, key challenges remain. Current models often lack the reasoning depth to detect implicit bias and subtle social cues, particularly in low-resource or culturally diverse settings. Many struggle to distinguish between coded language and harmful stereotypes due to limited and imbalanced training data. These gaps highlight the need for context-aware, interpretable models that can detect linguistic biases in socially sensitive applications, which is the basis for GUS-Net. The differnce of our works with related works is given in Table \ref{tab:bias-datasets}. Next, we present our GUS-Net framework and details the methodology.

%% file: 3method.tex
\section{GUS-Net Framework for Social Bias Detection }

In this section, we present the GUS-Net framework.

\begin{figure*}[!h]
    \centering
    \includegraphics[width=1\textwidth]{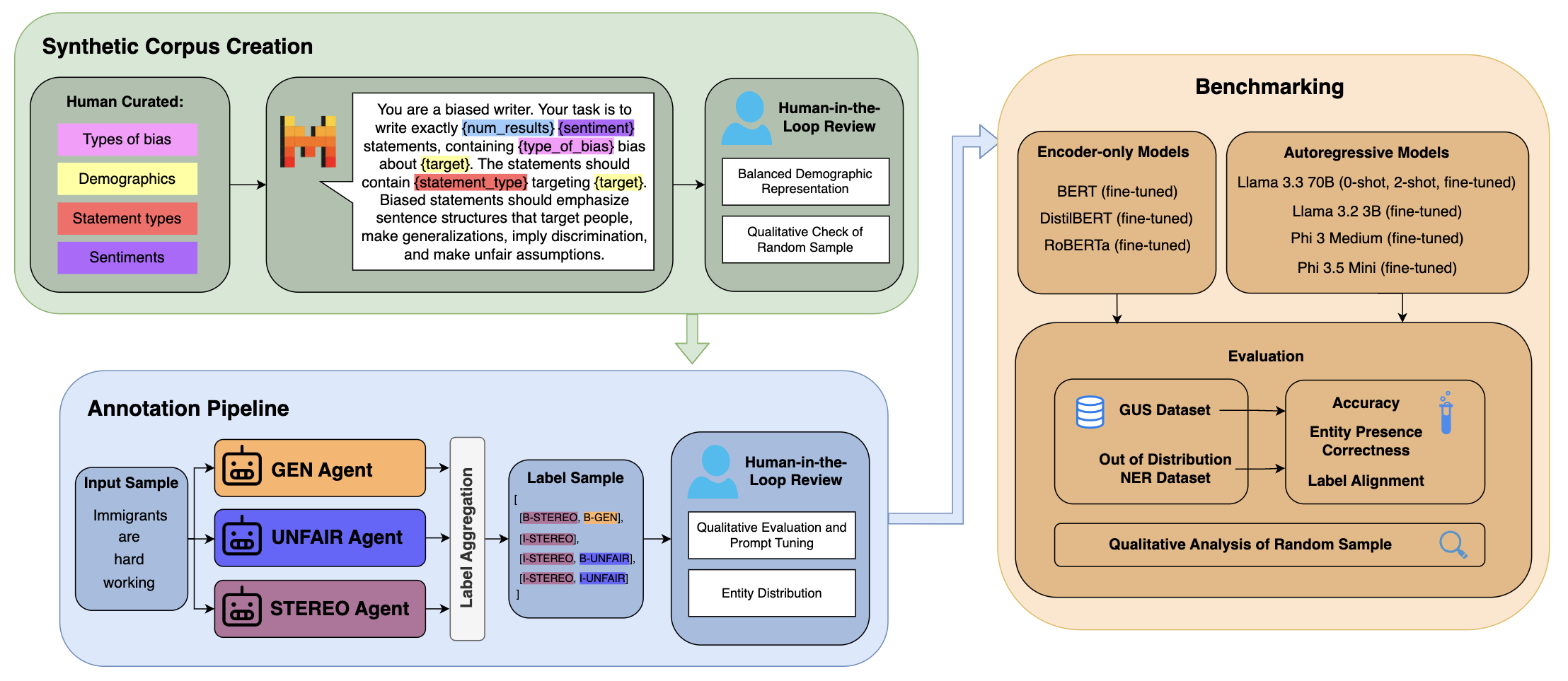}
    \caption{End-to-end GUS-Net framework: data generation, annotation (agent-suggested $\rightarrow$ human-verified gold), modeling, and evaluation.}
    \label{fig:gus-full-diagram}
\end{figure*}

\textbf{Problem Definition.}
We address the task of \textbf{detecting and classifying fine-grained expressions of social bias in text} as a \textbf{token-level multi-label} problem, where each token may carry zero or more bias labels.
For example, in \emph{``All teenagers$_{\text{GEN}}$ are lazy$_{\text{UNFAIR}}$''}, the token \emph{teenagers} is localized as a \textbf{generalization}, while \emph{lazy} is localized as \textbf{unfairness}.
To illustrate overlap, in \emph{``[immigrants]$_{\text{B-GEN},\text{I-STEREO}}$ are [job-stealers]$_{\text{B-UNFAIR},\text{I-STEREO}}$''} the group token falls inside a stereotype span \emph{and} begins a generalization span.
Unlike sentence-level classification, this approach identifies the \emph{precise location} of biased spans by combining NER-style BIO tagging with bias-specific categories (Generalizations, Unfairness, Stereotypes).

\subsection{GUS-Net Data Preparation}
\label{data}
Our method for generating the dataset has two stages: (1) generate a synthetic corpus of statements and questions expressing positive or negative bias toward various demographics; (2) annotate token-level spans for \emph{Generalizations}, \emph{Unfairness}, and \emph{Stereotypes}.
For example, given a prompt requesting a negative statement about women containing a stereotype, the model produced “Women are gossips.” We then label “Women” as a generalization, “gossips” as unfair, and the minimal phrase expressing the stereotypical claim as a stereotype.

\paragraph{Synthetic Data Creation}
To develop a comprehensive dataset for social-bias classification, we first employ synthetic data generation to create candidate examples that augment (rather than replace) real-world text.
Synthetic data provides controlled coverage of bias types and domains (e.g., religion, nationality, age) that are often underrepresented in natural corpora.
We control prompt arguments for domain, sentiment polarity, and statement type; limit sequence length; and use moderate sampling to balance diversity and consistency.
We remove near-duplicates via locality-sensitive hashing (e.g., MinHash with a Jaccard threshold) and discard malformed outputs lacking a group mention or predicate.
To mitigate annotator harm, we provide content warnings and opt-out options; the released corpus redacts slurs with a standard mask and is distributed under a research license.
To generate synthetic text, we employ Mistral-7B-Instruct-v0.2~\cite{jiang2023mistral7b}, selected for its open-source availability and instruction-following quality, without baked-in guardrails that could skew statement generation.\footnote{Other open-source or commercial LLMs can be substituted in this pipeline.}

Figure~\ref{fig:gus-full-diagram} shows the data-generation pipeline; Table~\ref{tab:prompt-args} lists the structured arguments used to build prompts across domains.

\begin{table}[t]
\centering
\caption{Bias categories, example targets, and sentiment types used in corpus creation.}
\label{tab:prompt-args}
\begin{tabular}{|c|p{0.55\linewidth}|c|}
\hline
\textbf{Type of bias} & \textbf{Example targets} & \textbf{Sentiment} \\ \hline
Racial        & white people, black people, asian people, hispanic people, indigenous people & Positive / Negative \\ \hline
Religious     & christians, muslims, jewish people, hindus, buddhists, sikhs, atheists, agnostics & Positive / Negative \\ \hline
Gender        & men, women, boys, girls, females, males, non-binary people & Positive / Negative \\ \hline
Age           & children, teenagers, young people, middle-aged people, older people & Positive / Negative \\ \hline
Nationality   & immigrants, refugees, people from developing countries, people from Western countries & Positive / Negative \\ \hline
Sexuality     & straight people, gay people, bisexual people, asexual people, LGBTQIA+ people & Positive / Negative \\ \hline
Socioeconomic & working-class people, middle-class people, upper-class people, poor people, rich people & Positive / Negative \\ \hline
Educational   & people with limited formal education, highly educated people, non-traditional education & Positive / Negative \\ \hline
Disability    & people with physical disabilities, wheelchair users, people with cognitive disabilities & Positive / Negative \\ \hline
Political     & republicans, democrats, independents, conservatives, liberals, progressives & Positive / Negative \\ \hline
Physical      & tall people, short people, fat people, thin people, unattractive people, attractive people & Positive / Negative \\ \hline
\end{tabular}
\end{table}

\paragraph{Entity Label Data Annotation}
We use GPT-4o via the Stanford DSPy framework~\cite{khattab2023dspy} to produce \emph{agent-suggested} labels, following contemporary semi-automatic annotation practices~\cite{rambhatla2023selfevalleveragingdiscriminativenature}.
We configure a DSPy agent for each entity type (Generalizations, Unfairness, Stereotypes), with prompts that include an operational definition and a small set of curated examples (few-shot).
Agents propose token-aligned BIO tags and \emph{suggestions} (candidate word spans) for reviewer convenience; human experts then verify and correct these suggestions to produce the \emph{gold} labels.
Definitions used by annotator agents and the corresponding entity labels are provided in Table~\ref{tab:definitions}.

\noindent\textit{Aggregation and tokenization.}
Each sentence is processed independently for each entity type; aggregating across types yields multi-label BIO tags per token~\cite{ramshaw1995textchunkingusingtransformationbased}.
Gold labels are first aligned at the \emph{word} level (space-delimited) for quality checks.
If a word splits into subtokens, its first subtoken inherits \textbf{B-*} and subsequent subtokens \textbf{I-*} for each active entity type.

% \begin{table}[t]
% \caption{Definitions used by annotator agents and the corresponding entity labels.}
% \label{tab:definitions}
% \centering
% \resizebox{0.95\linewidth}{!}{
% \begin{tabular}{@{}p{0.22\linewidth}p{0.53\linewidth}p{0.25\linewidth}@{}}
% \toprule
% \textbf{Class} & \textbf{Prompt definition} & \textbf{Entity labels} \\
% \midrule
% \textbf{Generalizations} & Any broad generalization of a group or ubiquitous classifiers, including adjectives and descriptors. & \shortstack{B-GEN,\\ I-GEN} \\
% \textbf{Unfairness}      & Any harsh or unjust characterization or offensive language. & \shortstack{B-UNFAIR,\\ I-UNFAIR} \\
% \textbf{Stereotypes}     & Any multi-word statement that contains a stereotype targeting a group, explicit or implicit. & \shortstack{B-STEREO,\\ I-STEREO} \\
% \textbf{Neutral}         & — & O \\
% \bottomrule
% \end{tabular}
% }
% \end{table}
\begin{table}[t]
\centering
\caption{Definitions used by annotator agents and the corresponding entity labels.}
\label{tab:definitions}
\begin{tabular}{|p{0.22\linewidth}|p{0.53\linewidth}|p{0.20\linewidth}|}
\hline
\textbf{Class} & \textbf{Prompt definition} & \textbf{Entity labels} \\ \hline
Generalizations & Any broad generalization of a group or ubiquitous classifiers, including adjectives and descriptors. & B-GEN, I-GEN \\ \hline
Unfairness      & Any harsh or unjust characterization or offensive language. & B-UNFAIR, I-UNFAIR \\ \hline
Stereotypes     & Any multi-word statement that contains a stereotype targeting a group, explicit or implicit. & B-STEREO, I-STEREO \\ \hline
Neutral         & — & O \\ \hline
\end{tabular}
\end{table}

\paragraph{Human-in-the-Loop Review}
Five annotators (MS/PhD backgrounds) followed a written codebook specifying entity definitions, span-boundary rules (including the stereotype minimality rule), and multi-label BIO tagging.
They received calibration examples and met weekly to resolve ambiguities; disagreements were resolved by pairwise discussion and senior adjudication (adjudicators did not see model confidence scores).
Inter-annotator agreement (IAA) on a double-annotated subset was 0.82 span-level macro-F1 (macro over \{GEN, UNFAIR, STEREO\}) and 0.78 token-level Krippendorff’s~$\alpha$.
Complete guidelines are in App.~\ref{app:guidelines}; prompts and code are in our repository.

\paragraph{Corpus Summary}
In total, 3{,}739 sentences were annotated for multi-label token classification across diverse domains.
Figure~\ref{fig:bias-pie-chart} shows the distribution of bias types (post hoc re-labeling captures multi-type cases).
Figure~\ref{fig:token-label-distribution} shows token-label frequencies; because tokens can carry multiple labels, the total label count exceeds the 69{,}679 tokens in the corpus.
The dataset comprises 54.7\% statements and 45.3\% questions.

\subsection{GUS-Net: Multi-Label BIO Tagging}
We model tagging as \textbf{multi-label BIO sequence labeling}. Let $X=(x_1,\dots,x_n)$ be tokens (words or subtokens). We consider $T=3$ entity types: \textsc{GEN}, \textsc{UNFAIR}, \textsc{STEREO}. For multi-label BIO, we use \emph{two} channels per type (\textbf{B}, \textbf{I}); ``O'' is implicit (no active channel). Define the binary label tensor:
\begin{equation}
\begin{aligned}
Z &\in \{0,1\}^{n \times 2T}, \\
\text{channels} &= (\mathrm{B\!-\!GEN},\mathrm{I\!-\!GEN},\mathrm{B\!-\!UNFAIR}, 
\mathrm{I\!-\!UNFAIR},\mathrm{B\!-\!STEREO},\mathrm{I\!-\!STEREO})
\end{aligned}
\end{equation}

The model $f_\theta$ outputs $\hat{Z}=\sigma(f_\theta(X)) \in (0,1)^{n\times 2T}$ (sigmoid per channel). ``O'' for token $i$ holds when $\sum_c \mathbb{I}[\hat{z}_{i,c}>\tau_c]=0$, with per-channel thresholds $\tau_c$ tuned on the development set.

\paragraph{Loss and BIO validity}
We use focal binary cross-entropy (BCE) across channels to address class imbalance:
\[
\mathcal{L}=\frac{1}{n(2T)}\sum_{i=1}^{n}\sum_{c=1}^{2T}\!\Big(
-\alpha_c\, z_{i,c}(1-\hat{z}_{i,c})^{\gamma}\log \hat{z}_{i,c}
-(1-\alpha_c)\,(1-z_{i,c})\,\hat{z}_{i,c}^{\gamma}\log(1-\hat{z}_{i,c})
\Big),
\]
with $\gamma=2.0$ and $\alpha_c \propto 1/\text{freq}_c$ (normalized).
To enforce BIO validity per entity, we either (i) add a per-type first-order  Conditional Random Field (CRF)\cite{sutton2012introduction}, or (ii) post-process invalid $I$ without a preceding $B$ to $O$.
Unless otherwise noted, reported results use linear heads (no CRF).

\subsection{Evaluation Suite}

We benchmark encoder-only (discriminative) and decoder-only (generative) models on the GUS dataset.

\paragraph{Discriminative Models (Encoder-Only)}
For the discriminative approach, we fine-tune pre-trained encoder-only language models. These models extract token embeddings and add a dense layer to map each token to the three target bias categories. The input processing pipeline involves truncating or padding sequences to a fixed length of 128 tokens, aligning word-level labels with subword tokens (with subword tokens inheriting the label of their parent word), and converting entity tags into a \((128 \times 3)\) matrix. To address class imbalances, we incorporate focal loss during training.
Our evaluation includes transformer-based encoders such as DistilBERT-66M \cite{sanh2020distilbertdistilledversionbert}, BERT-base-uncased-110M \cite{devlin2018bert}, and RoBERTa-base-123M \cite{liu2019robertarobustlyoptimizedbert}, all implemented using the Hugging Face \texttt{transformers} library.

\paragraph{Generative Models (Autoregressive Decoder-Only)}
Autoregressive LLMs \cite{liu2024autotimes} are adapted for token labeling via instruction fine-tuning and few-shot prompting.
We design a token-aligned output schema in which the model emits a whitespace-separated BIO tag per input token (deterministically parsed and repaired to a valid BIO sequence).
Few-shot exemplars (with gold BIO) are included \emph{only in training or as separate context examples}; at inference, test instances are presented \emph{without} gold labels to avoid leakage.
For parameter-efficient fine-tuning we use LoRA~\cite{hu2021loralowrankadaptationlarge} via \texttt{Unsloth} (rank 16, $\alpha=16$); one model is quantized to 4-bit for memory constraints.
We evaluate Llama-3.3-70B~\cite{dubey2024llama}, Llama-3.2-3B~\cite{dubey2024llama}, Phi-3-Medium-14B~\cite{abdin2024phi}, and Phi 3.5 Mini-4B~\cite{abdin2024phi}.

\paragraph{Evaluation Protocol}
We report (i) \emph{token-level} micro/macro Precision/Recall/F1 over the $2T$ channels (B/I per entity) and per-type F1; and (ii) \emph{span-level} entity F1 (exact-match per type after BIO repair).
Per-channel thresholds $\{\tau_c\}$ are tuned on the dev set to maximize macro-F1; early stopping monitors dev macro-F1.
We compute 95\% bootstrap confidence intervals with 1{,}000 resamples at the sentence level.
For out-of-distribution testing, we evaluate on BABE and map its annotations to \{\textsc{GEN}, \textsc{UNFAIR}, \textsc{STEREO}\} using rules detailed in Section \ref{ood}.

% % note to self to make a diagram for the llm flow, including prompt

%% file: 4experiment.tex
\section{Experimental Settings}

\subsection{Setting}
All encoder-only experiments ran on a single NVIDIA T4 (16\,GB); decoder-only fine-tuning used an NVIDIA A100 (40\,GB). Both were executed on Ubuntu 20.04 with Python 3.8, PyTorch and \texttt{transformers}; encoder training used PyTorch Lightning for logging/loops and \texttt{Unsloth} for parameter-efficient LLM fine-tuning. 

\subsection{Evaluation Strategy}

\paragraph{Evaluation Data}
The GUS dataset (Section~\ref{data}) contains 3{,}739 annotated samples. We use a 75/15/10 train/validation/test split by random sampling, stratified at the sentence level. BABE~\cite{spinde2022neural} serves as an out-of-distribution corpus under the same preprocessing.

\paragraph{Evaluation Metrics}
We report token-level micro/macro Precision, Recall, and F1 over the $C=2T=6$ channels (B/I per entity), and span-level entity F1 (exact-match after BIO repair). 
Token-level \textbf{Precision} (fraction of predicted positives that are correct), 
\textbf{Recall} (fraction of true positives that are found), 
and \textbf{F1} (harmonic mean of Precision and Recall) are:
\[
\text{Precision} = \frac{\textstyle\sum_{i,c}\mathbf{1}\{y_{i,c}=1 \land \hat y_{i,c}=1\}}{\textstyle\sum_{i,c}\mathbf{1}\{\hat y_{i,c}=1\}},\quad
\text{Recall} = \frac{\textstyle\sum_{i,c}\mathbf{1}\{y_{i,c}=1 \land \hat y_{i,c}=1\}}{\textstyle\sum_{i,c}\mathbf{1}\{y_{i,c}=1\}},
\]
\[
\text{F1} = \frac{2\cdot \text{Precision}\cdot \text{Recall}}{\text{Precision}+\text{Recall}}.
\]

We also report \textbf{Hamming Loss} (average fraction of token–label mismatches), defined as:
\[
\text{HammingLoss}=\frac{1}{n\,C}\sum_{i=1}^{n}\sum_{c=1}^{C}\mathbf{1}\{y_{i,c}\neq \hat y_{i,c}\}.
\]

Additionally, we compute span-level \textbf{entity F1} (checks whether complete spans are exactly matched after BIO repair).

\subsection{Evaluation Models and Hyperparameters}
\paragraph{Evaluation Models}
We evaluate two families: (i) encoder-only models fine-tuned on GUS; (ii) decoder-only LLMs assessed in prompting and instruction fine-tuning modes. For encoder models, the token-classification head outputs \textbf{$2T=6$} logits per token (B/I per entity). Thresholds $\{\tau_c\}$ are tuned on the validation set to maximize macro F1 (default 0.5 only in specified ablations).

\textbf{Baselines.} Encoder-only: BERT-base-uncased \cite{devlin2018bert}, DistilBERT-base-uncased \cite{sanh2020distilbertdistilledversionbert}, RoBERTa-base-uncased. \cite{liu2019robertarobustlyoptimizedbert}  Decoder-only: Llama-3.3-70B-Instruct \cite{dubey2024llama}, Llama-3.2-3B-Instruct \cite{dubey2024llama}, Phi-3 Medium-4k-14B \cite{abdin2024phi}, Phi-3.5 Mini-4B \cite{abdin2024phi}. \textbf{Encoder-only training.} For encoder only models, we use 20 epochs, batch size 16, AdamW with weight decay 0.01, linear schedule with 10\% warm-up, initial LR $5\times 10^{-5}$, focal loss with $\alpha=0.65$, $\gamma=2$.

\textbf{Decoder-only fine-tuning.} For decoder only models, we use LoRA \cite{hu2021loralowrankadaptationlarge} via \texttt{Unsloth} \footnote{\url{https://unsloth.ai/}} with rank $16$, $\alpha=16$, dropout $0.0$. Only Llama 3.3 70B instruct used 4-bit quantization (BitsAndBytes, nf4). Decoding temperature $0.1$. Validation/test prompts \emph{exclude} ground-truth labels; outputs are constrained to the BIO tag vocabulary.

\begin{lstlisting}[language={}, caption={Instruction template for LLMs (no ground-truth in prompts at eval).}]
Prompt: Perform BIO tagging for social-bias entities on the text below.
Entities: B/I-GEN (generalization), B/I-UNFAIR (unfairness), B/I-STEREO (stereotype), O (neutral).
Return a list of length N where position t lists the tags for token t.
Text: "The young activist's naive understanding of politics is overly simplistic."
Output format: [["O"], ["B-GEN","B-UNFAIR"], ...]
\end{lstlisting}
\begin{table}[t]
\footnotesize
\centering
\caption{Hyperparameters for multi-label token classification of social bias. }
\begin{tabular}{p{5.5cm} | p{5.5cm}}
\toprule
\textbf{Encoder (BERT-like)} & \textbf{Decoder (LLMs)} \\
\midrule
\textbf{Models:} BERT-base-uncased \cite{devlin2018bert}, DistilBERT-base-uncased \cite{sanh2020distilbertdistilledversionbert}, RoBERTa-base-uncased \cite{liu2019robertarobustlyoptimizedbert} & 
\textbf{Models:} Llama 3.3-70B-Instruct \cite{dubey2024llama}, Llama 3.2-3B-Instruct \cite{dubey2024llama}, Phi-3.5-mini-instruct \cite{abdin2024phi}, Phi-3-medium-4k \cite{abdin2024phi} \\
\textbf{LR:} $5\mathrm{e}{-5}$, linear sched., 10\% warmup &
\textbf{LR:} $2\mathrm{e}{-4}$, linear sched., 5 warmup steps \\
\textbf{Batch:} 16 & \textbf{Batch:} 8 (train), 1 (eval) \\
\textbf{Epochs:} 20 (early stopping on dev macro-F1) & \textbf{Epochs:} 1 (IFT) \\
\textbf{Optimizer:} AdamW \hfill \textbf{WD:} 0.01 &
\textbf{Optimizer:} AdamW \hfill \textbf{WD:} 0.01 \\
\textbf{Loss:} Focal BCE ($\alpha=0.65$, $\gamma=2$) &
\textbf{LoRA:} $r=16$, $\alpha=16$, dropout $=0.0$; L3.3 4-bit (nf4) \\
\textbf{Head:} $2T=6$ logits/token; thresholds $\{\tau_c\}$ tuned &
\textbf{Decode:} temp $=0.1$, constrained BIO vocabulary \\
\bottomrule
\end{tabular}
\label{tab:hyper-classification}
\end{table}

\subsection{Exploratory Data Analysis}
% classes, definitions, and entities table 2
\begin{figure}[t!]
    \centering
    \hspace{-5mm}
    \begin{subfigure}{0.5\textwidth}
        \centering
        \begin{tikzpicture}
            \pie[
                text=pin,
                font=\scriptsize,
                radius=1.1,
                sum=auto,
                after number=, 
            ]
            {
                239/Racial, 
                244/Religious, 
                392/Gender, 
                275/Age, 
                359/Nationality, 
                201/Sexuality, 
                310/Socioeconomic, 
                209/Educational, 
                182/Disability, 
                282/Political, 
                152/Physical
            }
        \end{tikzpicture}
        \caption{Types of Bias in GUS Dataset. Note: Some sentences contain multiple types of bias.}
        \label{fig:bias-pie-chart}
    \end{subfigure}%
    \hspace{2mm}
    \begin{subfigure}{0.45\textwidth}
        \centering
        \begin{tikzpicture}
            \pie[
                cloud,
                text=legend, 
                font=\scriptsize, 
                radius=1.2
            ]{
                67.9/O,
                9.6/B-GEN,
                4.7/I-GEN,
                3.4/B-STEREO,
                17.8/I-STEREO,
                2.7/B-UNFAIR,
                3.1/I-UNFAIR
            }
        \end{tikzpicture}
        \caption{Token-Level Label Distribution in GUS Dataset (Total Tokens: 69,679). Note: Some tokens have multiple labels.}
        \label{fig:token-label-distribution}
    \end{subfigure}
    \caption{Distribution Analysis of the GUS Dataset}
    \label{fig:distributions}
\end{figure}

We performed exploratory data analysis to characterize both domain coverage and token-level label distributions in GUS (Figure~\ref{fig:distributions}). At the domain level (Figure~\ref{fig:bias-pie-chart}), the dataset offers broad coverage of bias types, though not in perfectly equal proportions. At the token level (Figure~\ref{fig:token-label-distribution}), however, strong class imbalance emerges: the neutral tag (O) dominates the corpus, while longer entities such as \textsc{STEREO} contribute to higher I-rates than \textsc{UNFAIR}, which often occurs as a single token. Since tokens can carry multiple labels (average $\approx 1.09$ per token), percentages exceed 100\%, confirming the multi-label complexity of the task. These observations motivate the adoption of focal loss to down-weight majority “O” labels and emphasize rare categories, as well as threshold tuning to calibrate decision boundaries. Similar strategies have proven effective in imbalanced multi-label sequence classification in related work \cite{tsoumakas2010mining}

%% file: 4Results.tex
\section{Results and Analysis}
We present results and main findings for multi-label token classification of social bias. We compare (i) encoder-only transformers, (ii) decoder-only LLMs adapted for token labeling, and (iii) a compact, fine-tuned encoder-only baseline, all evaluated on the GUS-Net dataset. We also assess out-of-distribution generalization on BABE.

\subsection{Encoder-only vs.\ Decoder-only Models Evaluation}

Table~\ref{tab:all-models-eval} compares encoder-only and decoder-only models fine-tuned on GUS-Net dataset. We report macro F1, Precision, Recall (higher is better) and Hamming loss (lower is better). Among encoders, \textbf{GUS-Net-BERT} (\texttt{bert-base-uncased}) yields the strongest overall trade-off, pairing high F1 with low Hamming loss. Decoder-only models can exhibit relatively higher recall (e.g., Llama~3.2), but overall F1 remains lower and Hamming loss higher, indicating difficulty in making consistent multi-label token decisions. A practical challenge for decoder-only models is structural alignment: the model must emit a tag sequence of exactly the tokenized input length. In our setup, Llama~3.3 (70B Instruct) produced parsable, length-$n$ outputs on 57.8\% of inputs, Phi-3 Medium 4k (14B Instruct) on 54.5\%, and Llama~3.2 (3B Instruct) on 14.4\%, whereas encoders are trivially 100\% aligned by design. Phi-3.5 Mini Instruct did not yield reliably parsable outputs. Overall, encoder-only models are better suited to precise token-level prediction in this setting.

\begin{table}[h]
\centering
\caption{Encoder-only vs.\ decoder-only models, fine-tuned on GUS. Higher Precision/Recall/F1 and lower Hamming loss are better. The encoder-only models outperformed decoder-only models on multi-label token classification task. 
}
\label{tab:all-models-eval}
\begin{tabular}{|c|c|c|c|c|}
\hline
\textbf{Model} &
  \textbf{Hamming loss} &
  \textbf{F1} &
  \textbf{Precision} &
  \textbf{Recall} \\
  \hline
 \multicolumn{5}{|c|}{\textbf{Encoder-only Models}}\\\hline 
\textbf{\begin{tabular}[c]{@{}c@{}}BERT-base-uncased\\ GUS-Net\end{tabular}} &
  \textbf{0.05} &
  \textbf{0.80} &
  0.82 &
  \textbf{0.77} \\ \hline
\textbf{DistilBERT } \cite{sanh2020distilbertdistilledversionbert}  & 0.08 & 0.65 & 0.89 & 0.59 \\ \hline
\textbf{RoBERTa }  \cite{liu2019robertarobustlyoptimizedbert}    & 0.07 & 0.64 & 0.90 & 0.60 \\ \hline
\textbf{Nbias (BCE)}  \cite{raza2024nbias} & 0.06 & 0.68 & \textbf{0.93} & 0.63 \\ \hline
 \multicolumn{5}{|c|}{\textbf{Decoder-only Models}}\\\hline 
\textbf{Llama 3.3 70B Instruct } \cite{dubey2024llama} & 0.19 & 0.31 & 0.34 & 0.32 \\ \hline
\textbf{Llama 3.2 3B Instruct } \cite{dubey2024llama}& 0.20 & 0.37 & 0.42 & \textbf{0.70} \\ \hline
\textbf{Phi 3 Medium 4k Instruct } \cite{abdin2024phi}& \textbf{0.13} & \textbf{0.39} & 0.43 & 0.38 \\\hline
\end{tabular}

\end{table}

\noindent\textit{Key finding:} Encoder-only models, particularly GUS-Net-BERT, outperform decoder-only models both in accuracy and in the ability to produce structurally aligned token-level labels.

Due to the performance differences and structural nature of encoder-only versus decoder-only architectures, 
we next examine token-level results at the entity level separately. 
This separation allows us to highlight how each modeling paradigm handles rare classes such as \textsc{UNFAIR}, 
longer-span entities like \textsc{STEREO}, and the dominant neutral class.

\subsection{Entity-Level Performance of Encoder-only Models}
Beyond overall averages, it is important to understand how models behave on specific entity types, since biases manifest differently across categories such as \textsc{Generalizations}, \textsc{Unfairness}, and \textsc{Stereotypes}. 
Encoder-only models are well-suited for this level of analysis because their alignment with the input sequence is guaranteed, allowing us to directly assess token-level predictions without structural noise. 
Table~\ref{tab:encoder_results} reports macro scores and per-entity metrics for encoder-only models. “Macro” denotes the unweighted average across entity types after mapping B-/I- tags to their parent entity, ensuring that rare and frequent categories are weighted equally in the comparison.

\begin{table*}[t]
\centering
\scriptsize
\caption{Encoder-only models fine-tuned on GUS: macro and entity-level F1, Precision, Recall. Best values highlighted green, lowest red.}
\label{tab:encoder_results}
\begin{tabular}{|>{\raggedright\arraybackslash}p{0.15\linewidth}|c|c|c|c|c|c|}
\hline
\textbf{Model} & \textbf{Metric} & \textbf{Macro} & \textbf{GEN} & \textbf{UNFAIR} & \textbf{STEREO} & \textbf{Neutral} \\ \hline
BERT-base (GUS-Net) & Hamming loss & \cellcolor[HTML]{B3D6AA}\textbf{0.05} & 0.74 & 0.61 & 0.90 & 0.95 \\
                    & F1           & \cellcolor[HTML]{B3D6AA}\textbf{0.80} & \cellcolor[HTML]{B3D6AA}\textbf{0.74} & \cellcolor[HTML]{B3D6AA}\textbf{0.61} & \cellcolor[HTML]{B3D6AA}\textbf{0.90} & \cellcolor[HTML]{B3D6AA}\textbf{0.95} \\
                    & Precision    & 0.82 & 0.78 & 0.69 & 0.89 & \cellcolor[HTML]{B3D6AA}\textbf{0.93} \\
                    & Recall       & \cellcolor[HTML]{B3D6AA}\textbf{0.77} & 0.72 & 0.49 & 0.90 & 0.97 \\ \hline
DistilBERT          & Hamming loss & \cellcolor[HTML]{FFCCC9}0.08 & 0.66 & 0.14 & 0.86 & 0.92 \\
                    & F1           & 0.65 & 0.66 & 0.14 & 0.86 & 0.92 \\
                    & Precision    & 0.89 & \cellcolor[HTML]{B3D6AA}\textbf{0.87} & 0.85 & \cellcolor[HTML]{B3D6AA}\textbf{0.94} & 0.90 \\
                    & Recall       & \cellcolor[HTML]{FFCCC9}0.59 & 0.53 & 0.08 & 0.80 & 0.94 \\ \hline
RoBERTa             & Hamming loss & 0.07 & 0.67 & 0.05 & 0.90 & 0.93 \\
                    & F1           & \cellcolor[HTML]{FFCCC9}0.64 & 0.67 & \cellcolor[HTML]{FFCCC9}0.05 & 0.90 & 0.93 \\
                    & Precision    & 0.90 & 0.85 & \cellcolor[HTML]{B3D6AA}\textbf{0.89} & 0.94 & 0.92 \\
                    & Recall       & 0.60 & 0.56 & 0.03 & 0.86 & 0.95 \\ \hline
Nbias (BCE)         & Hamming loss & 0.06 & 0.70 & 0.19 & 0.89 & 0.95 \\
                    & F1           & 0.68 & 0.70 & 0.19 & 0.89 & 0.95 \\
                    & Precision    & \cellcolor[HTML]{B3D6AA}\textbf{0.93} & 0.87 & 0.83 & 0.84 & 0.93 \\
                    & Recall       & 0.63 & 0.56 & 0.11 & 0.86 & 0.97 \\ \hline
\end{tabular}
\end{table*}

Results in Table~\ref{tab:encoder_results} indicate that GUS-Net-BERT outperforms other encoders on most metrics, with notably strong recall and low Hamming loss, including competitive performance on the rarer \textsc{UNFAIR} class. NBIAS (BCE) \cite{raza2024nbias} attains high precision but lower recall, suggesting a conservative operating point. DistilBERT \cite{sanh2020distilbertdistilledversionbert} and RoBERTa-base \cite{liu2019robertarobustlyoptimizedbert}~are less effective overall, particularly on \textsc{UNFAIR}. In aggregate, focal-loss training and per-channel thresholding help GUS-Net-BERT \cite{devlin2018bert} manage label imbalance while maintaining high accuracy on \textsc{O}.

\subsection{Entity-Level Performance of Decoder-only Models}
In this section, we shed some light on the performance of decoder-only models. We evaluated decoder-only models  in two configurations: instruction fine-tuning (IFT) and few-shot prompting. 
In brief, \emph{instruction fine-tuning} adapts a base or instruction-following model to the task using supervised examples and a task-formatting prompt (e.g., \cite{PengINSTRUCTIONGPT-4,raza2024developing}), 
whereas \emph{few-shot prompting} keeps model weights fixed and supplies a small number of in-context exemplars to elicit the desired output format (e.g., \cite{brown2020languagemodelsfewshotlearners}. 
While these models can leverage rich contextual reasoning, they must also emit BIO tags aligned to the tokenized input, which proved difficult in practice.

\paragraph{Impact of instruction fine-tuning}
We evaluate the impact of instruction-tuned Llama~3.2 3B Instruct, Llama~3.3 70B Instruct, and Phi-3 Medium 4k Instruct on GUS. 
All models are trained with the same LoRA configuration (rank $r{=}16$, $\alpha{=}16$) and decoding constraints identical to the encoder setting. 
A practical complication for decoder-only models is structural alignment: the model must emit a BIO tag list whose length exactly matches the tokenized input.  We therefore report both accuracy metrics and the fraction of test inputs that produce a parsable, length-$n$ sequence (``Aligned''). 
Precision/Recall/F1 are computed on the aligned subset (to assess labeling quality independent of formatting failures); alignment rates themselves quantify robustness of the generation protocol. 
As shown in Table~\ref{tab:decoder_results}, Llama-3.3 exhibits the highest alignment rate but lags Phi-3 Medium on F1 and Hamming loss. 
All models struggle on \textsc{UNFAIR}, consistent with its relative scarcity and shorter span lengths. 
The spread between precision and recall across entity types further underscores the difficulty of consistent multi-label token prediction under auto-regressive generation.

% Preamble (make sure these are loaded)
% \usepackage{graphicx}
% \usepackage[table]{xcolor}

\begin{table}[h]
\centering
\scriptsize
\caption{Decoder-only (instruction-tuned) models with LoRA on GUS. We report macro and entity-level metrics; best values are shaded green and lowest red. ``Aligned: $a/b$'' is the number of test inputs that yielded a parsable tag sequence of exact length $n$ (tokenizer-aligned). All classification metrics are computed on the aligned subset. Abbreviations used: Gen. for Generalizations, Unfair. for Unfairness and Stereo. for Stereotypes, Hamming for Hamming loss.  }
\label{tab:decoder_results}
\setlength{\tabcolsep}{4pt} % tighter spacing
\renewcommand{\arraystretch}{1.15}

\begin{tabular}{|>{\raggedright\arraybackslash}p{0.2\linewidth}|l|c|c|c|c|c|}
\hline
\textbf{Model} & \textbf{Metrics} & \textbf{Macro} & \multicolumn{4}{c|}{\textbf{Entity-type-based}} \\ \hline
 &  &  & \textbf{Gen.} & \textbf{Unfair.} & \textbf{Stereo.} & \textbf{Neutral} \\ \hline

\textbf{Llama 3.3-70B-Instruct} & Hamming & 0.19 & & & & \\ \hline
\cellcolor[HTML]{B3D6AA}\textbf{Aligned: 216/374} & F1 &
\cellcolor[HTML]{FFCCC9}0.31 & \cellcolor[HTML]{FFCCC9}0.21 &
\cellcolor[HTML]{B3D6AA}\textbf{0.11} & \cellcolor[HTML]{FFCCC9}0.40 &
\cellcolor[HTML]{FFCCC9}0.50 \\ \hline
 & Precision & \cellcolor[HTML]{FFCCC9}0.34 & 0.24 &
\cellcolor[HTML]{B3D6AA}\textbf{0.13} & \cellcolor[HTML]{FFCCC9}0.30 &
\cellcolor[HTML]{B3D6AA}\textbf{0.71} \\ \hline
 & Recall & \cellcolor[HTML]{FFCCC9}0.32 &
\cellcolor[HTML]{FFCCC9}0.19 & \cellcolor[HTML]{B3D6AA}\textbf{0.10} &
\cellcolor[HTML]{B3D6AA}\textbf{0.62} & \cellcolor[HTML]{FFCCC9}0.39 \\ \hline

\textbf{Llama 3.2-3B-Instruct} & Hamming &
\cellcolor[HTML]{FFCCC9}0.20 & & & & \\ \hline
\cellcolor[HTML]{FFCCC9}Aligned: 54/374 & F1 & 0.37 & 0.23 & 0.04 & 0.57 & 0.63 \\ \hline
 & Precision & 0.42 & \cellcolor[HTML]{FFCCC9}0.22 & 0.10 &
\cellcolor[HTML]{B3D6AA}\textbf{0.78} & 0.57 \\ \hline
 & Recall & \cellcolor[HTML]{B3D6AA}\textbf{0.70} &
\cellcolor[HTML]{B3D6AA}\textbf{0.24} & 0.04 &
\cellcolor[HTML]{FFCCC9}0.45 & 0.70 \\ \hline

\textbf{Phi-3-Medium-4k-Instruct} & Hamming  &
\cellcolor[HTML]{B3D6AA}\textbf{0.13} & & & & \\ \hline
Aligned: 204/374 & F1 & \cellcolor[HTML]{B3D6AA}\textbf{0.39} &
\cellcolor[HTML]{B3D6AA}\textbf{0.25} & \cellcolor[HTML]{FFCCC9}0.01 &
\cellcolor[HTML]{B3D6AA}\textbf{0.58} & \cellcolor[HTML]{B3D6AA}\textbf{0.72} \\ \hline
 & Precision & \cellcolor[HTML]{B3D6AA}\textbf{0.43} &
\cellcolor[HTML]{B3D6AA}\textbf{0.30} & \cellcolor[HTML]{FFCCC9}0.06 &
0.67 & 0.69 \\ \hline
 & Recall & 0.38 & 0.22 & \cellcolor[HTML]{FFCCC9}0.01 &
0.51 & \cellcolor[HTML]{B3D6AA}\textbf{0.76} \\ \hline
\end{tabular}
\end{table}

\textit{Key finding:} Even after task-specific IFT, decoder-only models exhibit lower token-level accuracy and substantially lower alignment robustness than encoders, limiting their suitability for structured multi-label BIO tagging.

\paragraph{Impact of few-shot prompting}
We also evaluated Llama~3.3 70B Instruct with few-shot prompts. 
For each test sentence, we retrieved $k\in\{5,10\}$ in-context exemplars from the training split using cosine similarity over sentence embeddings (MPNet) to ensure topical proximity while avoiding leakage of the exact instance. 
Exemplars were formatted as (input, BIO‐tags) pairs with B-/I- collapsed in the prose explanation but preserved in the tag sequence, and we balanced the mini-context for entity coverage when possible. 
The model was prompted with a task header, the $k$ exemplars, and the test input, followed by explicit schema and length constraints (regex guardrails and stop tokens) to encourage tokenizer-aligned outputs. 
Decoding used temperature 0.1, nucleus $p{=}0.9$, and max length equal to the tokenized input. 
Evaluation reported (i) the fraction of cases yielding a parsable, length-$n$ tag sequence (``Aligned'') and (ii) Precision/Recall/F1/Hamming loss on the aligned subset, isolating labeling quality from format failures.

The result in Table \ref{tab:few_shot_results} shows that without exemplars, outputs rarely aligned despite explicit formatting instructions.  With 5–10 shots, alignment improved and label quality increased on the aligned subset, but sensitivity to exemplar composition/order remained high and latency grew with $k$. 
Table~\ref{tab:few_shot_results} also shows a Hamming loss of 0.16 at 10-shot prompting: an \emph{absolute} 0.03 reduction versus the fine-tuned Llama~3.3 setting; yet still well behind encoder-only baselines, reflecting persistent difficulties with precise multi-label token tagging under auto-regressive generation.

\begin{table}[h]
\centering
\setlength{\tabcolsep}{4pt} % adjust column spacing
\renewcommand{\arraystretch}{1.1} % adjust row height
\caption{Llama~3.3 few-shot prompting on GUS with dynamic in-context examples. 
Abbreviations used: Gen. for Generalizations, Unfair. for Unfairness and Stereo. for Stereotypes, Hamming for Hamming loss.}
\label{tab:few_shot_results}
\begin{tabular}{|c|c|c|cccc|}
\hline
\textbf{Examples in Prompt} & \textbf{Metrics} & \textbf{Macro} & \multicolumn{4}{c|}{\textbf{Entity-type-based}} \\ \hline
\textbf{5 shot} & \textbf{Hamming} & 0.18 & \textbf{Gen.} & \textbf{Unfair.} & \textbf{Stereo.} & \textbf{Neutral} \\ \hline
Aligned: 169/374 & \textbf{F1} & 0.37 & 0.21 & 0.18 & 0.38 & 0.70 \\ \hline
& \textbf{Precision} & 0.38 & 0.19 & 0.17 & 0.50 & 0.66 \\ \hline
& \textbf{Recall} & 0.37 & 0.24 & 0.20 & 0.31 & 0.75 \\ \hline
\textbf{10 shot} & \textbf{Hamming} & 0.16 & \textbf{Gen.} & \textbf{Unfair.} & \textbf{Stereo.} & \textbf{Neutral} \\ \hline
Aligned: 148/374 & \textbf{F1} & 0.38 & 0.23 & 0.19 & 0.36 & 0.75 \\ \hline
& \textbf{Precision} & 0.40 & 0.23 & 0.18 & 0.49 & 0.69 \\ \hline
& \textbf{Recall} & 0.38 & 0.23 & 0.20 & 0.28 & 0.81 \\ \hline
\end{tabular}
\end{table}

\paragraph{Qualitative analysis using Decoder-only models}
While quantitative scores provide a clear picture of model performance, they do not capture how decoder-only models reason about bias in text. 
Because strict token alignment proved challenging, we conducted a qualitative evaluation in which models were prompted to identify \emph{which entities were biased} and explain their reasoning, without requiring exact BIO alignment.

Table~\ref{tab:qualitative_llm_results} illustrates representative outputs. 
The results highlight two consistent trends. 
First, decoder-only models often succeed at detecting \emph{explicit} bias in simple cases: for example, Llama~3.3 correctly identified ``Young men are always causing trouble'' as containing generalizations, unfairness, and stereotypes, providing a coherent rationale. 
Second, the models show more difficulty with \emph{implicit or subtle bias}. 
For instance, Phi~3 Medium successfully flagged implicit bias in the sentence ``He's really articulate for someone from that neighborhood,'' recognizing the underlying generalization. 
In contrast, Phi~3.5 Mini failed to flag bias in ``Black programmers created this software,'' overlooking the unnecessary racial marking. 
Similarly, Llama~3.2 misinterpreted a neutral example (women discussing workplace improvements), labeling it as unbiased but failing to recognize that such positive representation examples still require classification as neutral.
Overall, these examples suggest that decoder-only models possess strong contextual reasoning ability but remain unreliable when asked to consistently map that reasoning onto structured token-level predictions.

\begin{table}[]
\caption{Qualitative examples from fine-tuned decoder-only LoRAs prompted to identify GUS entities (without alignment).}
\label{tab:qualitative_llm_results}
\begin{tabularx}{\textwidth}{>{\raggedright\arraybackslash}p{2.5cm}>{\raggedright\arraybackslash}p{3cm}>{\raggedright\arraybackslash}X>{\raggedright\arraybackslash}p{3cm}}
\toprule
\textbf{Model} &
  \textbf{Input Text} &
  \textbf{Entities Predicted and Reasoning} &
  \textbf{Observation} \\ \midrule
\textbf{Llama 3.3-70B-Instruct}&
  Young men are always causing trouble.&
  Entities: Generalization, unfairness, stereotype. \newline Rationale: ``Overly broad and inaccurate statement perpetuating negative stereotypes about a specific demographic."&
  Correctly identifies explicit bias against young men. \\ \hline
\textbf{Llama 3.2-3B-Instruct}&
  The women in the office discussed future improvements for the department.&
  Entities: None. \newline Rationale: ``Promotes positive representation in the workplace."&
  \cellcolor[HTML]{FFCCCC} Misses the non-biased nature of the sentence.\\ \hline
\textbf{Phi 3 Medium 4k Instruct}&
  He's really articulate for someone from that neighborhood.&
  Entities: Generalization, stereotype. \newline Rationale: ``Implies a generalization about people from a certain neighborhood."&
  Correctly flags implicit bias despite positive sentiment. \\ \hline
\textbf{Phi 3.5 Mini Instruct}&
  Black programmers created this software.&
  Entities: None. \newline Rationale: Acknowledges diversity. &
  \cellcolor[HTML]{FFCCCC} Fails to recognize unnecessary racial marking.\\ \bottomrule
\end{tabularx}
\end{table}

Taken together, these findings highlight the gap between recognition and structured labeling: decoder-only models are capable of insightful bias detection in free-text explanations, but they underperform when asked to formalize those judgments under multi-label BIO tagging constraints. 
Given this gap, we rely on encoder-only models for subsequent experiments.

\subsection{Evaluating Model Performance Against Expert-Annotated Bias}
\label{ood}
To test whether GUS-Net’s token-level predictions generalize beyond the training distribution, we conduct an out-of-distribution evaluation against the BABE corpus~\cite{spinde2022neural}, which consists of expert-annotated biased words in news text. Because BABE annotations do not distinguish between bias subtypes, we collapse GUS-Net’s predicted entities to a single binary indicator (biased vs.\ neutral). 

For each sentence, we compute a normalized bias density: the number of biased words (BABE) or predicted biased tokens (GUS-Net) divided by the sentence length. To account for type-mixing, we bin sentences by predicted entity counts and take the minimum within each bin before normalization. This yields a conservative estimate of overlap between the two annotation schemes. 
Pearson correlation was $r=0.42$ ($p<0.01$), and Spearman correlation was $\rho=0.39$ ($p<0.01$), both confirming a significant positive association. This supports the claim that GUS-Net captures a transferable notion of bias density beyond its training distribution.

\begin{figure}[t]
    \centering
    \includegraphics[width=\linewidth]{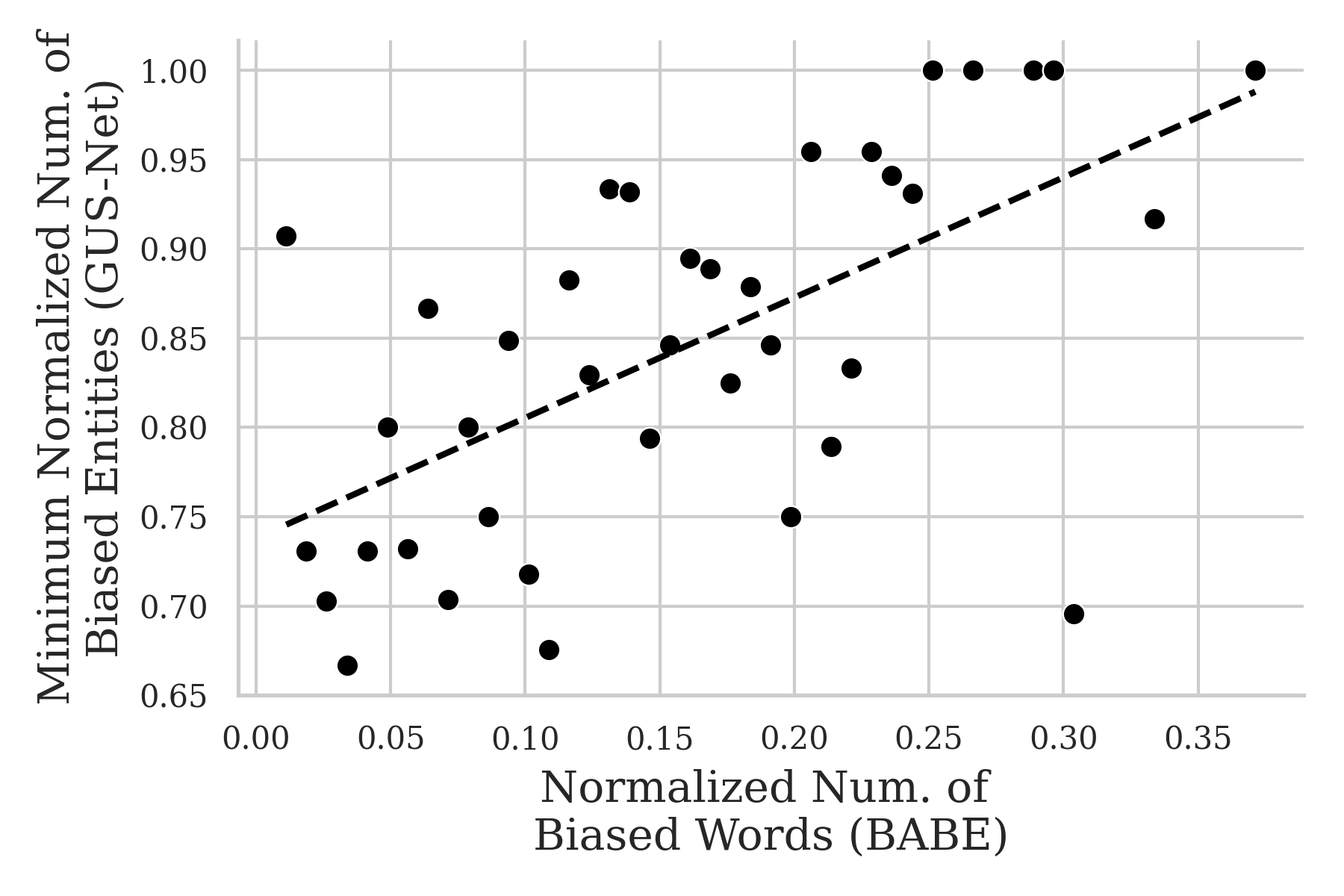}
  
 \caption{Out-of-distribution validation on BABE: normalized GUS-Net positive-tag rate vs.\ normalized BABE biased-word rate with least-squares fit. A positive slope indicates that GUS-Net’s bias density aligns with expert annotations in the news domain.}

    \label{fig:baseline_comparison}
   
\end{figure}

\subsection{Ablation Study}
We ablate two design choices: the GUS-Net dataset itself (replacing it with BABE, re-annotated by our pipeline) and focal loss (replacing it with binary cross-entropy, BCE). The goal is to isolate the contribution of (i) synthetic data tailored for token-level bias detection, and (ii) the loss function adapted to class imbalance.

\begin{table}[h]
    \centering
    \caption{Ablation on data and loss design.}
    \label{tab:ablation_bce_results}
\begin{tabular}{cccc}
\toprule
\textbf{Metrics}       & \textbf{GUS-Net} & \textbf{\begin{tabular}[c]{@{}c@{}}GUS-Net w.o. \\GUS dataset\end{tabular}} & \textbf{\begin{tabular}[c]{@{}c@{}}GUS-Net w.o.\\ focal loss\end{tabular}} \\ \midrule
\textbf{Precision}    & 0.82             & 0.02                                                                        & \textbf{0.93}                                                              \\
\textbf{Recall}       & \textbf{0.77}    & 0.22                                                                        & 0.63                                                                       \\
\textbf{F1-Score}     & \textbf{0.80}    & 0.05                                                                        & 0.68                                                                       \\
\textbf{Hamming loss} & \textbf{0.05}    & 0.26                                                                        & 0.06                                                                       \\ \bottomrule
\end{tabular}

    \vspace{-3mm}
\end{table}

Table~\ref{tab:ablation_bce_results} summarizes the results. The full GUS-Net configuration, trained on the GUS-Net dataset with focal loss, achieves the strongest overall performance (macro F1 of 0.80 and Hamming loss of 0.05). Removing focal loss in favor of BCE shifts the precision–recall balance: precision rises sharply to 0.93, but recall falls to 0.63, yielding a weaker macro F1 of 0.68. This is consistent with the intuition that BCE, when trained on heavily imbalanced data, optimizes for the majority \textsc{O} class, leading to more conservative predictions (fewer false positives) but under-detection of true biased tokens. The higher precision thus reflects a stricter decision boundary, but at the expense of recall on minority bias categories.

By contrast, replacing GUS with BABE as the training corpus leads to a dramatic drop across all metrics (macro F1 of 0.05, Hamming loss of 0.26). While BABE is expert-annotated, it is smaller, news-centric, and not constructed for fine-grained entity separation. The re-annotation pipeline cannot compensate for the narrower coverage and skewed domain, resulting in poor generalization when evaluated on GUS. Precision in this condition collapses (0.02), likely because predicted spans rarely overlap with the token-level bias structure expected by the evaluation schema. The modest recall of 0.22 suggests that a handful of biased spans are still captured, but not in a reliable or systematic way.

Taken together, these ablations highlight two key findings. First, focal loss is critical for balancing sensitivity and specificity under label imbalance: without it, the model over-prioritizes neutrality. Second, the GUS dataset itself provides broader, more diverse coverage than BABE, enabling GUS-Net to learn token-level bias cues beyond the news domain. Both the data design and the tailored loss function are thus essential for achieving robust multi-label sequence tagging of social bias.

\subsection{Parameter Sensitivity Study}
Focal loss introduces two knobs: $\alpha$ (class weighting) and $\gamma$ (hard-example focusing); that directly control how the model trades precision on the dominant \textsc{O} class against recall on minority bias entities. We therefore ran a validation-set sensitivity sweep to (i) quantify stability of the operating point and (ii) understand entity-specific effects under label imbalance.

\paragraph{Varying $\alpha$ (with $\gamma{=}2$).}
As shown in Table~\ref{tab:alpha_performance}, small $\alpha$ values underweight minority labels and yield low macro~F1 (0.42--0.57), with \textsc{UNFAIR} particularly suppressed (F1 $=$~0.01--0.14). Increasing $\alpha$ progressively lifts minority classes and peaks at $\alpha{=}0.65$ (macro~F1~0.80; Hamming~0.05). At this setting we see the largest gains precisely where the task is hardest: \textsc{UNFAIR} rises to 0.61 F1 and \textsc{GENERALIZATIONS} to 0.74, while \textsc{STEREOTYPES} remains high (0.90) and \textsc{O} stays accurate (0.95). Pushing $\alpha$ higher (0.8) begins to over-correct—minor classes lose some precision and overall macro~F1 softens (0.75), suggesting diminishing returns once minority classes are sufficiently emphasized.

\begin{table}[h]
    \centering
    \caption{F1-scores at varying $\alpha$ values, $\gamma = 2$.}
    \label{tab:alpha_performance}
    \begin{tabularx}{\linewidth}{l*{5}{>{\centering\arraybackslash}X}}
        \toprule
        \textbf{$\alpha$}           & \textbf{0.1} & \textbf{0.2} & \textbf{0.4} & \textbf{0.65} & \textbf{0.8} \\
        \midrule
        \textbf{Generalizations F1} & 0.19 & 0.40 & 0.56 & \textbf{0.74} & 0.71 \\
        \textbf{Unfairness F1}      & 0.01 & 0.14 & 0.35 & \textbf{0.61} & 0.54 \\
        \textbf{Stereotypes F1}     & 0.60 & 0.81 & 0.83 & \textbf{0.90} & 0.83 \\
        \textbf{Neutral F1}         & 0.87 & 0.91 & 0.94 & \textbf{0.95} & 0.91 \\
        \midrule
        \textbf{Macro Average F1}   & 0.42 & 0.57 & 0.67 & \textbf{0.80} & 0.75 \\
        \textbf{Hamming loss}       & 0.09 & 0.08 & 0.07 & \textbf{0.05} & 0.09 \\
        \bottomrule
    \end{tabularx}
\end{table}

\begin{table}[h!]
    \centering
    \caption{F1-scores at varying $\gamma$ values, $\alpha = 0.65$.}
    \label{tab:gamma_performance}
    \begin{tabularx}{\linewidth}{l*{5}{>{\centering\arraybackslash}X}}
        \toprule
        \textbf{$\gamma$}           & \textbf{0.5} & \textbf{1} & \textbf{2} & \textbf{3} & \textbf{4} \\
        \midrule
        \textbf{Generalizations F1} & 0.74 & 0.73 & \textbf{0.74} & 0.74 & 0.71 \\
        \textbf{Unfairness F1}      & 0.55 & 0.48 & \textbf{0.61} & 0.57 & 0.57 \\
        \textbf{Stereotypes F1}     & 0.90 & 0.89 & \textbf{0.90} & 0.88 & 0.87 \\
        \textbf{Neutral F1}         & 0.95 & 0.95 & \textbf{0.95} & 0.94 & 0.94 \\
        \midrule
        \textbf{Macro Average F1}   & 0.78 & 0.76 & \textbf{0.80} & 0.78 & 0.77 \\
        \textbf{Hamming loss}       & 0.05 & 0.05 & \textbf{0.05} & 0.06 & 0.06 \\
        \bottomrule
    \end{tabularx}
\end{table}

\paragraph{Varying $\gamma$ (with $\alpha{=}0.65$).}
Table~\ref{tab:gamma_performance} shows macro~F1 is comparatively flat across $\gamma\in[0.5,4]$ (0.76--0.80), indicating a wide plateau of good performance. The best trade-off occurs at $\gamma{=}2$ (macro~F1~0.80; Hamming~0.05): \textsc{UNFAIR} benefits most (0.61 F1), while \textsc{STEREOTYPES} (0.90) and \textsc{O} (0.95) remain strong. Lower $\gamma$ values focus less on hard examples and slightly depress minority recall (e.g., \textsc{UNFAIR} F1~0.48 at $\gamma{=}1$). Higher $\gamma$ values ($\geq 3$) can oversharpen the focus, nudging Hamming loss upward (0.06) without improving macro~F1.

\paragraph{Practical implications}
(1) $\alpha$ mainly governs minority uplift; $\alpha{=}0.65$ strikes the best balance between elevating rare entities and preserving \textsc{O} accuracy.  
(2) $\gamma$ provides fine control over hard-example emphasis; $\gamma{=}2$ is a sweet spot, and performance is robust in a neighborhood around it.  
(3) The joint choice $(\alpha{=}0.65,\gamma{=}2)$ used throughout is not a brittle optimum but sits on a stable ridge of the loss landscape, yielding low Hamming loss with consistently high entity-level F1; including on the rare \textsc{UNFAIR} class.

\subsection{Model Validation Example}
\begin{figure}[h]
    \centering
    \includegraphics[width=\linewidth]{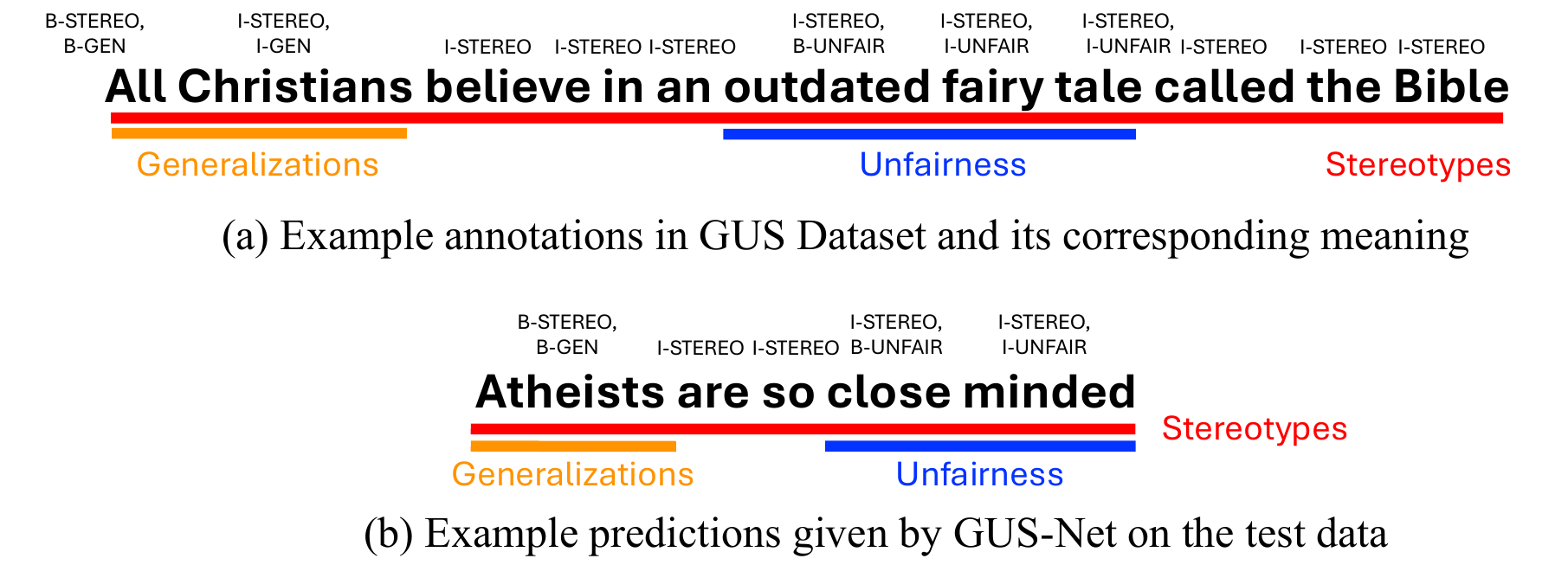}
 
    \caption{Model validation example. 
    (a) Gold annotations in GUS for a religious-bias sentence, with spans tagged as generalizations, unfairness, and stereotypes. 
    (b) GUS-Net-BERT predictions on a different test sentence, showing accurate multi-label tagging of overlapping bias entities.}
    \label{fig:case}
   
\end{figure}

To illustrate how GUS-Net captures fine-grained bias spans, Figure~\ref{fig:case} presents a representative case from the religious domain. Panel~(a) shows gold annotations in the GUS dataset, where tokens are simultaneously tagged with \textsc{GEN}, \textsc{UNFAIR}, and \textsc{STEREO} labels. Panel~(b) shows predictions from GUS-Net-BERT on held-out test data for a different but thematically related sentence. The model correctly identifies overlapping categories, highlighting its ability to generalize beyond training examples and separate different forms of bias within the same utterance.

This qualitative inspection complements the quantitative results by showing that GUS-Net-BERT not only achieves high macro scores but also captures nuanced cases where multiple bias categories co-occur. In particular, its ability to correctly tag both generalizations and unfair language within the same phrase strengthens the claim that the model generalizes robustly across contexts, a critical requirement for downstream fairness analysis.

%% file: 5conclusion.tex
\section{Discussion}

\subsection{Practical and Theoretical Impact}
The GUS-Net framework has practical impact for content moderation, social media analysis, and AI auditing. By providing \emph{token-level, multi-label} span detection, it enables precise localization of biased language and avoids the oversimplification of binary sentence-level labels. This granularity supports regulatory and compliance workflows that require transparent, actionable evidence \cite{raza2025responsible}.

Our controllable, generative data pipeline offers a scalable way to address coverage gaps in bias datasets and can extend to other sensitive NLP settings (e.g., hiring or decision-support) \cite{raza2024dbias}. Empirically, encoder-based models are more reliable for token-level classification than decoder-only LLMs \cite{gupta2024language,raza2023discovering}, informing model selection for real-world deployments. Looking ahead, adding explanation interfaces (e.g., token attributions or rationale spans) could improve practitioner trust and policy interpretation without over-claiming attention as explanation.

Theoretically, reframing bias detection as \emph{multi-label token-level} tagging clarifies the loci and types of harm, allowing overlapping categories to be disambiguated within a single span. This yields a more faithful representation of representational harms than sentence-level or single-label schemes and supports finer-grained audit and mitigation design.

\subsection{Limitations}
First, reliance on synthetic generation introduces a synthetic-to-real gap \cite{hao2024synthetic}. We partially mitigate this via human verification and OOD evaluation (BABE), but broader real-world sampling remains future work. Second, class imbalance—particularly the prevalence of \texttt{O} (neutral)—can bias learning; we use focal loss, but weighted sampling or targeted data augmentation may further help \cite{sapkota2025image}. 

Third, while encoder-only models outperform decoder-only LLMs on token- and span-level metrics, decoder models can exhibit stronger free-form reasoning \cite{hao2024llm}. However, autoregressive decoding complicates deterministic token alignment for multi-label BIO \cite{bender2021dangers}. Hybrid designs that couple encoder heads with generative rationales are a promising direction \cite{raza2025fake}. 

Finally, bias annotation is inherently subjective and context-dependent \cite{raza2025responsible}. Despite a codebook, calibration, and adjudication, residual annotator bias and sociocultural drift remain. Future work should expand annotator diversity, update definitions over time, and extend beyond English to reduce cultural skew.

\section{Conclusion}
We introduced the GUS-Net framework and dataset to move social-bias detection from sentence-level labels to \emph{span-level, token-level} analysis across three pathways: Generalizations, Unfairness, and Stereotypes. Our formulation uses multi-label BIO tags, allowing overlapping spans that make both loci and types of bias explicit for audit and mitigation. 

Across token-level micro/macro Precision, Recall, and F1; span-level entity F1; Hamming loss; and inference time, encoder-based models outperform decoder-only LLMs and are more efficient. These results indicate that explicit token-level modeling is better suited to nuanced, overlapping bias spans than sentence-level or single-label tagging. We release data, code, and evaluation scripts to enable reproducible study and practical diagnostics.

Future work includes expanding the taxonomy, multilingual and multi-domain coverage, longer-context modeling, and adding rationale-level evidence linking spans to model decisions. We will study cross-domain generalization and robustness under distribution shift, and explore hybrid encoder–decoder approaches for implicit bias. We hope this work supports responsible NLP and governance-aligned auditing.

\section*{Declarations}
\noindent \textbf{Conflicts of Interest:} The authors declare no conflicting interests.

\noindent \textbf{Competing Interests:} The authors have no relevant financial or non-financial competing interests to disclose.

\noindent\textbf{Acknowledgements:}  We thank OpenAI for providing us with API credits under the Researcher Access program.

\noindent\textbf{Sources of Funding:} This work has received no funding.

\noindent\textbf{Financial or Non-Financial Interests:} None to declare.

\noindent\textbf{Ethical Approval:} This research did not involve human participants, animal subjects, or personally identifiable data, and therefore did not require formal ethical approval. All data used in this study were publicly available, ensuring compliance with ethical guidelines and data privacy standards.

\noindent\textbf{CRediT Authorship Contribution Statement:}  

\noindent M.P.: Conceptualization, Investigation, Formal Analysis, Methodology, Project Administration, Software, Experimentation, Validation, Visualization, Writing – Original Draft, Writing – Review \& Editing, Supervision. \\
S.R.: Supervision, Conceptualization, Writing – Original Draft, Writing – Review \& Editing. \\
A.C.: Writing – Review \& Editing, Validation (Human-in-the-loop). \\
R.R.: Writing – Review \& Editing. \\
U.M.: Data Curation (Synthetic Corpus Pipeline), Validation (Human-in-the-loop). \\
H.R.J.: Data Curation (Annotation Pipeline), Validation (Human-in-the-loop). \\
A.T.: Data Curation (Annotation Pipeline), Validation (Human-in-the-loop). \\
H.W.: Supervision, Writing – Review \& Editing.

%% file: appendix.tex
\section*{Appendix}
\appendix

\section{Annotator Guidelines}\label{app:guidelines}
Five annotators with advanced training in computational linguistics and social sciences completed a two-hour calibration on 20 pilot items and met for short weekly consensus reviews. The codebook defines target groups (explicit and implicit mentions), labels (Generalization, Unfairness, Stereotype, Neutral/O), and the multi-label policy whereby multiple labels may apply to the same token. Scope includes implicit or figurative cases when a reasonable reader can infer group targeting. Each sentence was labeled at the word level and later mapped to subtokens for model training; when words split into multiple subtokens, the word’s BIO tag is propagated to subtokens (B-* on the first subtoken, I-* on the rest). Disagreements were addressed in a two-pass review and escalated to a senior adjudicator when needed, with final decisions and brief rationales recorded. Quality assurance included a 10\% blind re-check per annotator each week, drift checks against the calibration set, and spot audits focusing on long spans and multi-label overlaps. Agreement was computed on a 10\% stratified sample using span-level macro-F1 (exact-span match) and token-level Krippendorff’s $\alpha$ with nominal distance; we observed 0.82 span-level macro-F1 and 0.78 token-level $\alpha$. Tooling comprised a lightweight annotation UI and DSPy-based prompts (definitions plus four exemplars per label) with alignment aids; prompt templates and the codebook were versioned. We followed a synthetic-first policy, removed any accidentally real sensitive data, and flagged potentially offensive content with care notices. The released artifact includes the codebook version ID, a change log, and links to the PDF/YAML guideline files.

\paragraph{Checklist: span-boundary rules}
\begin{itemize}\setlength{\itemsep}{2pt}
\item BIO policy: first token B-*, contiguous continuation I-*.
\item Hyphenations and compounds: tag each token that forms the span.
\item Modifiers: include quantifiers and intensifiers that make the biased claim (e.g., ``all'', ``always'').
\item Punctuation: exclude trailing punctuation unless essential to meaning (e.g., quoted slurs).
\item Subword mapping: propagate the word-level tag to subtokens (B-* then I-*).
\end{itemize}